\documentclass[conference]{IEEEtran}
\ifCLASSINFOpdf
  \usepackage[pdftex]{graphicx}
  \graphicspath{./figs/}
  \DeclareGraphicsExtensions{.pdf,.jpeg,.JPG,.jpg,.png,.PNG}
\else
\fi
\usepackage{amsmath}
\usepackage{algorithmic}

\usepackage{array}

\ifCLASSOPTIONcompsoc
  \usepackage[caption=false,font=normalsize,labelfont=sf,textfont=sf]{subfig}
\else
  \usepackage[caption=false,font=footnotesize]{subfig}
\fi
\usepackage{url}

\usepackage{color}
\usepackage{algorithm}

\definecolor{codeblue}{rgb}{0,0,1.0}
\definecolor{codegreen}{rgb}{0,0.6,0}
\definecolor{codegray}{rgb}{0.5,0.5,0.5}
\definecolor{codepurple}{rgb}{0.58,0,0.82}
\definecolor{backcolour}{rgb}{0.95,0.95,0.92}

\usepackage{listings}
\usepackage{framed,enumitem}
\usepackage{multirow}

\usepackage{tikz}
\usepackage{textcomp}
\usepackage{hyperref}
\usepackage{lipsum}

\newcommand\copyrighttext{%
  \footnotesize \textcopyright 2019 IEEE. Personal use of this material is permitted. Permission from IEEE must be obtained for all other uses, in any current or future media, including reprinting/republishing this material for advertising or promotional purposes, creating new collective works, for resale or redistribution to servers or lists, or reuse of any copyrighted component of this work in other works.}
\newcommand\copyrightnotice{%
\begin{tikzpicture}[remember picture,overlay] at (current page.south)
\node[anchor=south,yshift=10pt] at (current page.south) {\fbox{\parbox{\dimexpr\textwidth-\fboxsep-\fboxrule\relax}{\copyrighttext}}};
\end{tikzpicture}%
}

\begin{document}
%
\title{Interpreting OWL Complex Classes in AutomationML based on Bidirectional Translation}

\author{\IEEEauthorblockN{Yingbing Hua\IEEEauthorrefmark{1}, Bj\"orn Hein\IEEEauthorrefmark{2}}
\IEEEauthorblockA{Faculty of Informatics\\
Karlsruhe Institute of Technology\\
Karlsruhe, Germany\\
Email: \IEEEauthorrefmark{1}yingbing.hua@kit.edu, \IEEEauthorrefmark{2}bjoern.hein@kit.edu}
}


%


\maketitle

\copyrightnotice

\begin{abstract}
The World Wide Web Consortium (W3C) has published several recommendations for building and storing ontologies, including the most recent OWL 2 Web Ontology Language (OWL). These initiatives have been followed by practical implementations that popularize OWL in various domains. For example, OWL has been used for conceptual modeling in industrial engineering, and its reasoning facilities are used to provide a wealth of services, e.g. model diagnosis, automated code generation, and semantic integration. More specifically, recent studies have shown that OWL is well suited for harmonizing information of engineering tools stored as AutomationML (AML) files. However, OWL and its tools can be cumbersome for direct use by engineers such that an ontology expert is often required in practice. Although much attention has been paid in the literature to overcome this issue by transforming OWL ontologies from/to AML models automatically, dealing with OWL complex classes remains an open research question. In this paper, we introduce the \textit{AML concept models} for representing OWL complex classes in AutomationML, and present algorithms for the bidirectional translation between OWL complex classes and their corresponding AML concept models.
We show that this approach provides an efficient and intuitive interface for non-experts to visualize, modify, and create OWL complex classes.
\end{abstract}


%
\IEEEpeerreviewmaketitle

\section{Introduction}

The World Wide Web Consortium (W3C) has published several recommendations for building ontologies, with the Resource Description Framework (RDF) and the Web Ontology Language (OWL) being the most popular ones. OWL was designed as an extension of RDF with significant more expressivity and is preferred as a language for conceptual modeling in complex domains. The reasoning facilities of OWL can, therefore, be used to support decision making in the domain of interest.

The Automation Markup Language (AutomationML, or AML) is a neutral, XML-based data format for data exchange between engineering tools \cite{Rainer2008}. AML is standardized as IEC 62714 and has its root in the data format CAEX (IEC 62424). AML supports the modeling of plant topology, component structure, geometry and kinematics, logic behavior, and communication networks. However, AML per se does not provide a formal semantics for automated data interpretation \cite{Hua2018a}. In practice, tools need to achieve a common understanding of the data and be responsible for the preservation of semantics.

Efforts have been made on adopting OWL and its reasoning facilities for processing AML data. The typical approach comprises three steps: a) transform engineering data stored in an AML document to an AML ontology by explicitly define the semantics of AML notions; b) after communication with the domain experts, an ontology expert extends the AML ontology with additional knowledge for specific engineering purposes; c) utilizing the reasoner for providing advanced engineering services. For example, with predefined ontological descriptions about error types in plant models, Abele et al. were able to identify modeling errors in the plant topology \cite{Abele2013a}. Hua et al. proposed a model-driven robot programming approach that is capable of inferring component capability and the associated programming interfaces from AML models \cite{Hua2016}. In this paper, we use the term \textit{AML ontology} to indicate an OWL ontology that is converted from an AML document.

It is evident that the approaches mentioned above are based on sophisticated domain knowledge that is modeled as OWL complex classes by nesting logic-based OWL constructors.
Therefore, a profound understanding of the domain and the language OWL is required. Recently, Hildebrandt et al. proposed the \textit{domain expert-centric} approach for building ontologies of cyber-physical systems \cite{Hildebrandt2018}. While this approach tackles the problem of incorporating domain expert's knowledge, it is unclear yet how to deal with OWL complex classes. In the remainder of the paper, we use the term \textit{OWL complex class} and the term \textit{OWL class} interchangeably if the context is clear.

In this paper, we introduce the \textit{AML concept model} for representing ontological semantics in native AML models. Based on a bidirectional translation procedure between OWL and AML, OWL classes can be visualized as AML concept models for inspection and modification, and \textit{proper} AML concept models can be transformed to OWL classes while preserving the ontological semantics. We show that this approach demonstrates an efficient and intuitive interface for a non-expert to interact with OWL complex classes.

This paper is organized as follows. Section \ref{sec:related_work} discusses related work on model transformation between OWL and AML. Section \ref{sec:preliminaries} gives a brief overview of OWL and AML, and introduces the important notions used in this paper. In section \ref{sec:amlmodel} we present the \textit{AML concept model} that is developed for preserving ontological semantics in AML. In section \ref{sec:translate} we describe the bidirectional translation between OWL complex classes and AML concept models. Finally, we demonstrate the utility of this approach with two typical use cases of ontology engineering in section \ref{sec:usecase} and conclude the paper with future works in section \ref{sec:conclussion}.

\section{Related Work}
\label{sec:related_work}

The first result about converting AML to OWL appeared in 2009 by Runde et al. in their German paper \cite{Runde09}. Two approaches were proposed and discussed. The \textit{abstract approach} represents the CAEX vocabulary directly as OWL classes in the ontology and transforms CAEX classes, objects and attributes as individuals of these OWL classes. The \textit{concrete approach} generates an OWL class for each CAEX class with an annotation about its original type in the CAEX schema. For example, an AML role class $\mathsf{Robot}$ will be converted to an OWL class with the annotation $\mathrm{RoleClass}$. Subsequent researches generally follow either the abstract or the concrete approach. For example, Kovalenko et al. proposed a lightweight ontology for covering core concepts of CAEX using the abstract approach \cite{Kovalenko2015}, while Hua et al. followed the concrete approach for learning unknown engineering concepts from AML data \cite{Hua2018a}.

The backward transformation from OWL to AML is less studied, although the first approach was already published in 2010 in \cite{Runde10}. The transformation begins with mapping atomic OWL classes to appropriate CAEX classes using the CAEX type annotation of each OWL class. It proceeds with OWL individuals of the top level OWL classes and transforms them into proper CAEX objects. Then the transformation handles each property associated with the individuals until all information in OWL is processed.

It is evident that existing methods only target at "simple" knowledge types, that is, atomic classes, objects, and properties. For handling complex ontological knowledge, e.g. OWL complex classes, one challenge arises that no regular AML model can preserve complex ontological semantics.

In the remainder of the paper, we assume that we are given an AML ontology converted from an AML document following the approach proposed in \cite{Hua2018a}. Our goal is to develop a modeling approach that enhances native AML models with ontological semantics and a translation procedure between such native AML models and OWL complex classes.

\section{Preliminaries}
\label{sec:preliminaries}

\subsection{AutomationML}
\label{sec:aml}
AML data is stored in an XML document which conforms to the underlying CAEX XML schema. An AML document usually contains a set of class libraries and a structured collection of engineering objects that represents the plant topology. We emphasize the following core concepts of CAEX that we consider in this paper:

\begin{itemize}
    \item \textbf{Role class} ($\mathbf{RC}$): a role class refers to a type of engineering objects, e.g. $\mathsf{Robot}$. As AML follows the object-oriented paradigm, role classes can be organized in inheritance hierarchies within so-called \textit{role class libraries}.
    \item \textbf{Interface class} ($\mathbf{IC}$): an interface class represents a type of engineering interfaces, e.g. $\mathsf{SignalInterface}$ or $\mathsf{AttachmentInterface}$. Similar to role classes, inheritance is allowed between interface classes and an \textit{interface class library} stores a set of interface classes. 
    \item \textbf{Internal element} ($\mathbf{IE}$): an internal element is the model of an engineering object, e.g. a joint inside a robot or a real robot in the plant. By referring to a role class, the meaning of an internal element is declared. For describing the plant topology, internal elements are organized as tree structures in the \textit{instance hierarchy}.
    \item \textbf{External interface} ($\mathbf{EI}$): an external interface is the model of an engineering interface, e.g. an IO pin of a controller. The type of an external interface is defined by referring to an interface class.
    \item \textbf{System unit class} ($\mathbf{SUC}$): a system unit class is a reusable engineering template that contains an internal structure, where internal elements are used to represent individual parts of the structure. 
\end{itemize}

For all the concepts mentioned above, CAEX attributes can be defined to describe their properties. In the rest of the paper, we use the notion \textit{AML model} to refer to any XML model that can be generated according to the CAEX schema.

\subsection{OWL}
\label{sec:owl}

OWL\footnote{While OWL is the short name of the Web Ontology Language whose expressive power goes beyond the scope of description logics, we use this notion to refer to the specific sub-language OWL 2 DL.} belongs to the family of expressive Description Logics (DL) and is closely related to $\mathcal{SROIQ}$ \cite{OWL2DL}. An OWL ontology defines a finite set of classes (e.g. $\mathsf{Robot}$), individuals (e.g. a robot instance) and properties in a domain of discourse, and describes relations between these artifacts. One further distinguishes between \textit{object properties} and \textit{data properties}. The former one is used for relations between individuals (e.g. a robot has a controller), and the latter one is for describing the concrete qualities of an individual (e.g. weight of a robot). Although an AML ontology generated by \cite{Hua2018a} has merely two object properties $\mathsf{hasIE}/\mathsf{hasEI}$, we also consider the following inverse properties in this paper: $$\mathsf{isIEOf} \equiv \mathsf{hasIE^-}, \mathsf{isEIOf} \equiv \mathsf{hasEI^-}$$

An OWL class is either an atomic class or a complex one when it is generated by so-called concept constructors \cite{OWL2DL}. Table \ref{table:dl} shows the concept constructors of OWL, their correspondences in the terminology of DL, their DL syntax\footnote{Please refer to \cite{Markus2012} for more details of the DL syntax.}, and their formal model-theoretic semantics in OWL. We use conventional notions for the syntax: $A$ represents an atomic class, $C$ or $D$ stands for an OWL (complex) class, $R$ stands for an OWL property, $a$ or $b$ stands for an OWL individual, $DR$ is used for the data range of data properties, and $lt$ is used for a literal value. The nested OWL class $C$ inside a restriction e.g. $\exists R.C$ is called the \textit{filler} of the restriction. 

\begin{table*}[htbp]
\centering
\caption{Syntax and semantics of OWL constructors}
\label{table:dl}
\begin{tabular}{|c|c|c|c|c|}
\hline
& \textbf{OWL Terminology} & \textbf{DL Terminology}      & \textbf{DL Syntax}     & \textbf{Semantics} \\ \hline
\multirow{13}{*}{\shortstack{\textbf{covered} \\ in this paper}} & atomic class             & atomic concept               & $A$                 & $A^\mathcal{I}$ \\ 
& Thing                    & top concept                  & $\top$              & $\Delta^\mathcal{I}$ \\ 
& Nothing                  & bottom concept               & $\bot$              & $\emptyset$ \\ 
& ObjectIntersectionOf           & intersection                 & $C \sqcap D$        & $C^\mathcal{I} \cap D^\mathcal{I}$ \\ 
& ObjectUnionOf                  & union                        & $C \sqcup D$        & $C^\mathcal{I} \cup D^\mathcal{I}$ \\ 
& ObjectOneOf                    & nominal                      & $\{a, b, ...\}$             & $\{a^\mathcal{I}, b^\mathcal{I}, ...\}$ \\
& ObjectSomeValuesFrom           & existential restriction      & $\exists R.C$       & $\{x | \exists y.(x,y) \in R^\mathcal{I} \land y \in C^\mathcal{I}\}$ \\ 
& DataSomeValuesFrom           & existential restriction      & $\exists R.(DR)$       & $\{x | \exists y.(x,y) \in R^\mathcal{I} \land y \in (DR)^\mathcal{I}\}$ \\ 
&  ObjectExactCardinality           & exact restriction         & $= n R.C$        & $\{x \,|\,\, |\{\,y | (x,y) \in R^{\mathcal{I}} \land y \in C^{\mathcal{I}}\,\}| = n\}$ \\ 
& ObjectMinCardinality           & at-least restriction         & $\geq n R.C$        & $\{x \,|\,\, |\{\,y | (x,y) \in R^{\mathcal{I}} \land y \in C^{\mathcal{I}}\,\}| \geq n\}$ \\ 
& ObjectMaxCardinality           & at-most restriction          & $\leq n R.C$        & $\{x \,|\,\, |\{\,y | (x,y) \in R^{\mathcal{I}} \land y \in C^{\mathcal{I}}\,\}| \leq n\}$ \\ 
& ObjectHasValue                    & fills restriction                     & $\exists R.\{a\}$             & $\{x |(x, a^\mathcal{I}) \in R^\mathcal{I}\}$ \\
& DataHasValue                    & fills restriction                     & $\exists R.\{lt\}$             & $\{x |(x, (lt)^\mathcal{I}) \in R^\mathcal{I}\}$ \\ 
& ObjectAllValuesFrom            & universal restriction        & $\forall R.C$       & $\{x | \forall y.(x,y) \in R^\mathcal{I} \to y \in C^\mathcal{I}\}$ \\ 
& ObjectComplementOf             & complement                   & $\neg C$            & $\Delta^\mathcal{I} \backslash C^\mathcal{I}$ \\ \hline

\multirow{7}{*}{\shortstack{\textbf{uncovered} \\ in this paper}} 
& DataExactCardinality           & exact restriction         & $= n R.(DR)$        & $\{x \,|\,\, |\{\,y | (x,y) \in R^{\mathcal{I}} \land y \in (DR)^{\mathcal{I}}\,\}| = n\}$ \\ 
& DataAllValuesFrom            & universal restriction        & $\forall R.(DR)$       & $\{x | \forall y.(x,y) \in R^\mathcal{I} \to y \in (DR)^\mathcal{I}\}$ \\ 
& DataMinCardinality           & at-least restriction         & $\geq n R.(DR)$        & $\{x \,|\,\, |\{\,y | (x,y) \in R^{\mathcal{I}} \land y \in (DR)^{\mathcal{I}}\,\}| \geq n\}$ \\ 
& DataMaxCardinality           & at-most restriction          & $\leq n R.(DR)$        & $\{x \,|\,\, |\{\,y | (x,y) \in R^{\mathcal{I}} \land y \in (DR)^{\mathcal{I}}\,\}| \leq n\}$ \\ 
& ObjectHasSelf                  & local reflexivity            & $\exists R.Self$    & $\{x \,|\, (x,x) \in R^{\mathcal{I}}\}$ \\ \hline
\end{tabular}
\end{table*}

In this paper, we consider an OWL complex class constructed by using arbitrarily many of the constructors in the \textit{covered} part of Table \ref{table:dl}, and assume that no constructors in the \textit{uncovered} part would appear. While it seems to be a strong assumption, we argue that: 
a) since CAEX attributes are mapped to data properties and one CAEX attribute is usually assigned to an object only once, the universal, at-least and at-most restrictions on data properties can be omitted;
b) the local reflexivity cannot appear in an AML ontology, since we can not have an internal element (or external interface) which is the internal element (or external interface) of itself. 

Formula \ref{eq:classes} shows some examples of OWL classes used in this paper. 
Class A refers to $\mathsf{Robot}$s without any internal element of the type $\mathsf{\neg IOController}$. 
Class B refers to internal elements of a $\mathsf{Robot}$ from the manufacturer KUKA. 
Class C refers to $\mathsf{Robot}$s with an $\mathsf{IOController}$ that has at least three $\mathsf{IOInterface}$s. 
Class D refers to $\mathsf{IOInterface}$s from objects that have at least three $\mathsf{IOInterface}$s.
Apparently, as the complexity grows, the intended meaning of an OWL complex class becomes more difficult to understand. In the next section, we introduce the \textit{AML concept model} that is able to represent OWL complex classes as native AML models.
\begin{align}
&\mathsf{A} \equiv \mathsf{Robot \sqcap \neg \exists hasIE.(\neg IOController)} \nonumber \\ 
&\mathsf{B} \equiv \mathsf{\exists isIEOf.(Robot \sqcap hasManufacturer."KUKA"}) \nonumber \\
&\mathsf{C} \equiv \mathsf{Robot \sqcap \exists hasIE.(IOController \,\, \sqcap \geq 3 hasEI.IOInterface}) \nonumber \\
&\mathsf{D} \equiv \mathsf{IOInterface \sqcap \exists isEIOf.(\geq 3 hasEI.IOInterface}) 
\label{eq:classes}
\end{align}

\section{The AML Concept Model}
\label{sec:amlmodel}

Consider the OWL class constructors in Table \ref{table:dl}. It is evident that while atomic classes can be represented as AML classes directly \cite{Hua2018a}, most of the features in OWL are not supported by AML. Therefore, we propose the following approach to represent OWL class constructors as dedicated AML models:

\textbf{Atomic class}: similar to the conventional translation procedure as proposed by \cite{Runde10}, an atomic class is represented by a CAEX role or interface class. A class reference in CAEX is therefore equivalent to a class assertion in OWL. For example, an internal element $a$ of the role class $A$ is represented as $A(a)$.

\textbf{Thing}: Thing is the most general concept in OWL and contains all individuals. Therefore, it is represented by a CAEX object with no specific configurations.

\textbf{Nothing}: Nothing is the most specific concept in OWL and contains no individual. Nothing is handled as the complement of Thing (see the complement case below).

\textbf{Intersection}: an intersection $C \sqcap D$ contains individuals that are instances of all the operands $C$ and $D$ in the intersection. Therefore, an intersection is represented by the composition of several AML models that correspond to each of the operands, including CAEX class references, attributes, and subordinate object structures.

\textbf{Union}: a union $C \sqcup D$ contains individuals that are instances of at least one operand $C$ or $D$ of the union. XML does not support unions in general. In this paper, we handle each operand of a union separately and generate one AML model for each of them. 

\textbf{Nominal}: a nominal $\{a, b, ...\}$ enumerates all individuals that an OWL class shall contain. Similar to the union constructor, nominals cannot be directly represented in XML, and we generate one AML model for each element inside a nominal.

\textbf{Existential restriction}: an existential restriction $\exists R.C$ or $\exists R.(DR)$ states the existence of the relation $R$ with the filler $C$ or the data range $DR$. If $R$ is an object property, the existential restriction is represented by a child object (internal element or external interface) while the filler $C$ is represented by the model of the child object. If $R$ is a data property, the existential restriction is represented by a CAEX attribute while the data range $DR$ is represented by the configuration of the CAEX attribute, e.g. data type and value requirements.

\textbf{Object cardinality restrictions}: an object cardinality restriction, i.e. an exact restriction $= n R.C$, an at-least restriction $\geq n R.C$, or an at-most restriction $\leq n R.C$, defines the number of child objects of the class $C$ w.r.t. the relation $R$. The CAEX attributes \textit{minCardinality} and \textit{maxCardinality} are added to the child objects to represent the minimum and maximum number respectively. The exact cardinality of $n$ is represented by $minCardinality = minCardinality = n$.

\textbf{Fills restriction}: a fills restriction $\exists R.\{a\}$ or $\exists R.\{lt\}$ corresponds to an existential restriction with a Singleton filler. If $R$ is an object property, the CAEX attribute \textit{isIdentifiedByID} is used to restrict the ID of the child object, as ID is unique in AML. If $R$ is a data property, $lt$ is set as the required value of the corresponding CAEX attribute.

\textbf{Universal restriction}: a universal restriction $\forall R.C$ forces all child objects w.r.t. the relation $R$ to be instances of the class $C$. For example, $\mathsf{\forall hasIE.C}$ describes things that have internal elements of type $\mathsf{C}$ only. While universal restrictions can not be directly represented in XML, it can be simulated by disallowing child objects that are instances of the class $\neg C$ \cite{OWL2} using the exact cardinality $= 0 \, R.(\neg C)$.

\textbf{Complement}: a complement $\neg C$ contains all individuals that are not instances of $C$. Since an OWL class can have arbitrarily nested complements, we first transform an OWL class to its negation normal form (NNF) so that complements are only bound to atomic classes \cite{Hitzler:2009}. For example, the NNF of the OWL class A in Formula \ref{eq:classes} is: $$\mathsf{NNF(A) \equiv Robot \sqcap \forall hasIE.IOController}$$ Obviously, $\mathsf{NNF(A)}$ does not contain any complements. In fact, complements can only appear in the following three cases in the NNF of an OWL class:

\begin{itemize}
\item[(a)] A complement can be bound to an atomic class as $\neg A$ or a data range as $\neg DR$, and is not part of any restrictions. In this case, a CAEX attribute \textit{negated=true} is added to the AML model. Note that intersections of a mixture of positive and negative atomic classes, e.g. $\neg A_1 \sqcap A_2$, cannot be modeled in AML.
    
\item[(b)] A complement can be the filler of an existential restriction, i.e. $\exists R.(\neg A)$ or $\exists R.(\neg DR)$. As with the existential restriction, a child CAEX object or CAEX attribute is first generated. Then the CAEX attribute \textit{negated=true} is added to the child model.
    
\item[(c)] A complement can be the filler of a universal restriction as $\forall R.(\neg A)$ (recall that we ignore universal restrictions on data properties). In this case, we disallow child objects of the class $A$ w.r.t. the relation $R$, which can be expressed using the exact cardinality $= 0 \, R.A$.
\end{itemize}

Table \ref{table:attributes} summarizes the introduced CAEX attributes that are used to capture the semantics of OWL constructors mentioned above. We call them \textit{concept attributes}. The attribute \textit{primary} is a helper flag to indicate which element in an AML model is described by the OWL class. We call an AML model with concept attributes as an \textit{AML concept model} and enumerate the values of concept attributes based on possible forms of NNF in Table \ref{table:mapping}. Intuitively, AML concept models can be nested to represent nested OWL class expressions. An AML concept model is \textit{proper} if it has exactly one primary element. Note that intersections, unions, and nominals are omitted in the mapping since we handle each element of them individually. 

\begin{table*}[htbp]
\centering
\caption{The AML concept attributes for capturing ontological semantics.}
\label{table:attributes}
\begin{tabular}{|c|c|c|l|c|}
\hline
\textbf{Name}    & \textbf{Type} & \textbf{Default} & \multicolumn{1}{c|}{\textbf{Designation}}                              & \textbf{Semantics in OWL}       \\ \hline
negated          & bool               & false                  & whether the class reference or the data range of an AML concept model shall be negated              & complement                     \\ \hline
minCardinality   & integer            & 1                      & minimum number of occurrence of this AML concept model                 & minCardinality \\ \hline
maxCardinality   & integer            & unlimited                     & maximum number of occurrence of this AML concept model                & maxCardinality    \\ \hline
identifiedByID  & bool               & false                & whether the ID of the AML concept model is used as an individual name in OWL      & nominal \\ \hline
primary         & bool               & false                  & whether this AML concept model is the primary object    & target individuals \\ \hline
\end{tabular}
\end{table*}

\begin{table*}[htbp]
\centering
\caption{Mapping between OWL constructors and AML concept attributes.}
\label{table:mapping}
\begin{tabular}{|c|c|c|c|c|}
\hline
\textbf{OWL Class Expression} & \textbf{Negated} & \textbf{minCardinality} & \textbf{maxCardinality} \\ \hline
simple complement $\neg C$ or $\neg DR$ & true     & 1     & unlimited    \\ \hline
existential restriction $\exists R.C$ or $\exists R.DR$    & false     & 1     & unlimited    \\ \hline
existential restriction $\exists R.(\neg C)$ or $\exists R.(\neg DR)$  & true     & 1     & unlimited    \\ \hline
universal restriction  $\forall R.C$  & true     & 0     & 0    \\ \hline
universal restriction  $\forall R.\neg C$  & false     & 0     & 0    \\ \hline
at-least restriction $\geq n R.C$    & false     & n     & unlimited    \\ \hline
at-most restriction $\leq n R.C$   & false     & 0     & n    \\ \hline
\end{tabular}
\end{table*}

\begin{figure}[tbp]
\centering
\subfloat[Class A]{\includegraphics[width=3.4cm]{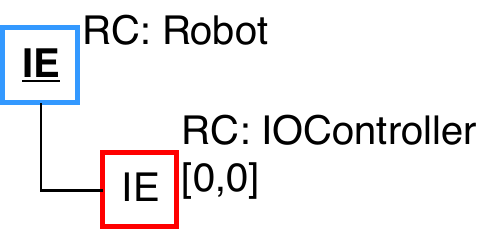}%
\label{fig:classA}}
\subfloat[Class B]{\includegraphics[width=3.35cm]{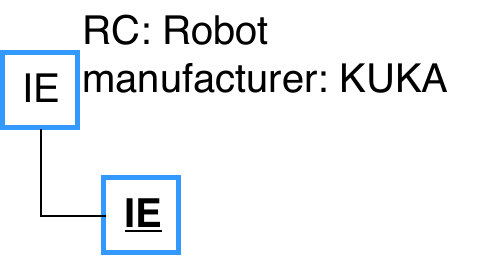}%
\label{fig:classB}}
\\
\subfloat[Class C]{\includegraphics[width=3.8cm]{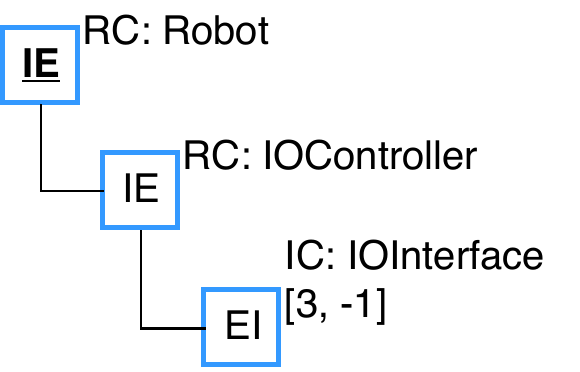}%
\label{fig:classC}}
\subfloat[Class D]{\includegraphics[width=3.37cm]{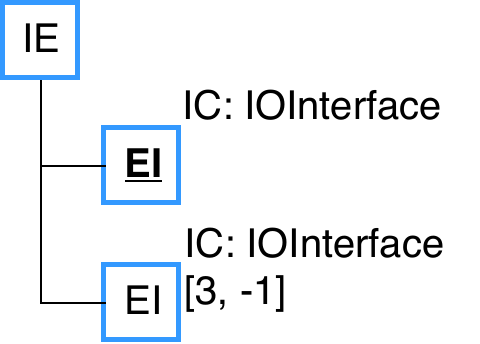}%
\label{fig:classD}}
\caption{The AML concept models for the OWL classes in Formula \ref{eq:classes}.}
\label{fig:classes}
\end{figure}

Figure \ref{fig:classes} illustrates the AML concept models of the NNF of the OWL classes A, B, C and D in Formula \ref{eq:classes} as tree structures. Internal elements (IE) and external interfaces (EI) are represented by tree nodes, and their class references and attributes are depicted as labels on the top right corner. A negated object is marked as red. The primary object is marked as bold with an underline. Numbers in square brackets are the min and max cardinality of the object, while a value $-1$ means that it is unlimited. Note that for the classes B and D, the primary object is not the root node since XML cannot depict "part-of" relations (i.e. $\mathsf{isIEOf, isEIOf}$). Therefore, each inverse property is simulated as a predecessor node in the XML tree.

\section{Translation between OWL and AML}
\label{sec:translate}

The core idea of the translation is to exploit the \textbf{tree structure} of OWL class expressions. More concretely, we introduce \textit{AML concept trees} that depict OWL complex classes in a tree structure similar to AML concept models. Then we describe the \textit{forward} translation $\mathrm{TransF: OWL \mapsto AML}$ via the AML concept trees. Finally, we show that the \textit{backward} translation $\mathrm{TransB: AML \mapsto OWL}$ can be directly carried out using the mappings in Table \ref{table:mapping}.

\subsection{From OWL to AND-tree}
\label{sec:owl2tree}

We define a tree conventionally as a directed graph $\mathcal{G = (V, E)}$ where $\mathcal{V}$ is a finite set of nodes and $\mathcal{E}$ is a finite set of edges, to which the following rules apply:
\begin{itemize}
    \item A tree $\mathcal{G}$ has a unique \textit{root} node that has no predecessor.
    \item Each node $n \in \mathcal{V}$ has a unique predecessor.
\end{itemize}

We call \textit{leaf nodes} the tree nodes that have no successor, i.e. at the bottom of the tree. Furthermore, a \textit{branching node} is an inner tree node that has a unique predecessor and arbitrarily many successors. Based on these notions, an \textit{AND-tree} is a tree with the following properties:
\begin{itemize}
    \item The root of an AND-tree represents the expression of an OWL complex class.
    \item Each branching node of an AND-tree represents either an intersection or a restriction (see the notions in Table \ref{table:dl}).
    \item Each leaf node of an AND-tree represents either OWL Thing, OWL Nothing or an atomic class.
\end{itemize}

For each OWL complex class without unions and inverse properties, an AND-tree can be constructed by making a successor node for each operand of an intersection and the filler of a restriction, as shown in Algorithm \ref{alg:and1}. 
\begin{algorithm}[tbp] 
\caption{$\mathrm{Construct}$} 
\label{alg:and1} 
\begin{algorithmic}[1] 
\renewcommand{\algorithmicrequire}{\textbf{Input:}}
\renewcommand{\algorithmicensure}{\textbf{Output:}}
\REQUIRE The class expression $ce$ of an OWL class $C$
\ENSURE  A tree node $root$

\STATE {make a tree node $root$ for $ce$}
\IF {$ce$ is an atomic class}
        \RETURN $root$
\ELSIF {($ce$ is an intersection)}
        \FOR {$operand \in ce$}
            \STATE {let $child = $ Construct($operand$)}
            \STATE {add $child$ as a successor to $root$}
        \ENDFOR
\ELSIF {($ce$ is a restriction)}
        \STATE {let $child = $ Construct($ce.filler$)}
        \STATE {add $child$ as a successor to $root$}
\ENDIF
\RETURN $root$
\end{algorithmic}
\end{algorithm}

\begin{figure}[htbp]
\centering
\includegraphics[width=0.4\textwidth]{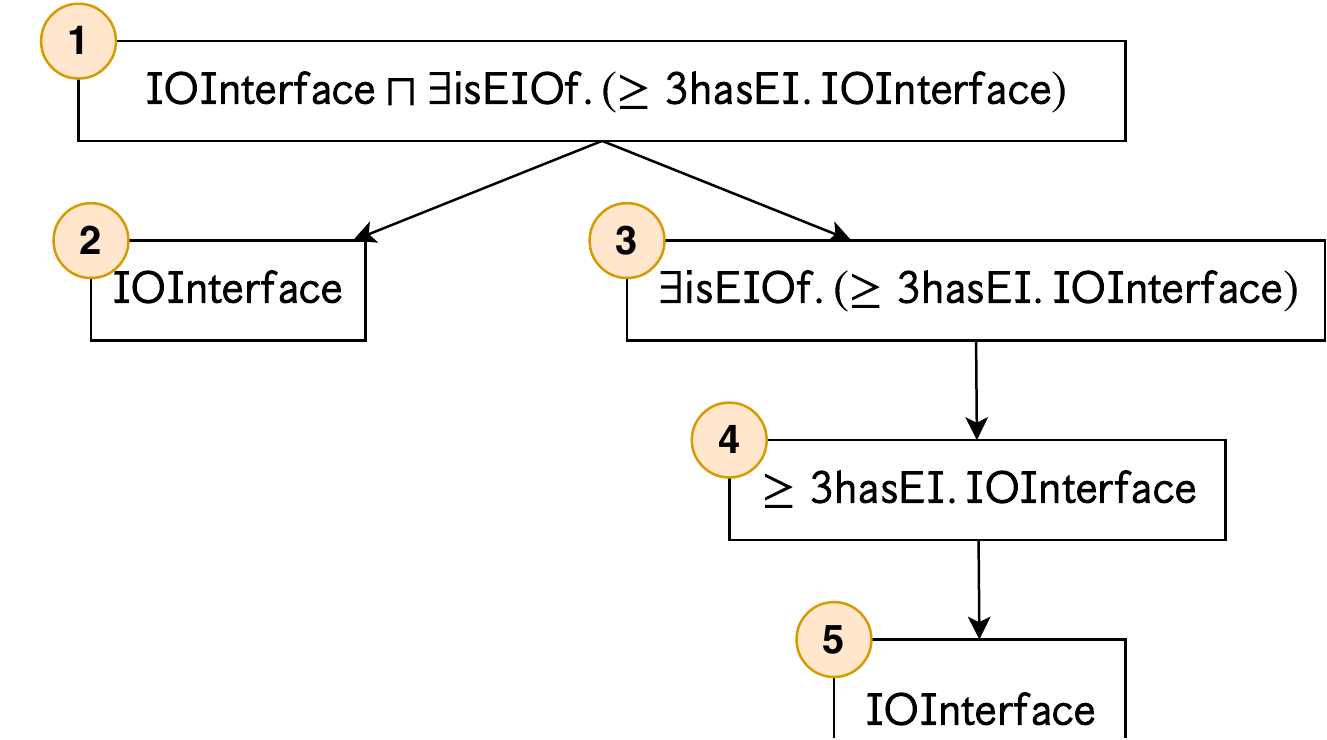}
\caption{The AND-tree constructed from the OWL class D in Formula \ref{eq:classes}. The numbers in the tree nodes show the sequence of node construction.}
\label{fig:andtree}
\end{figure}

We illustrate the construction process in Figure \ref{fig:andtree}. Each box represents a tree node, and the number on the upper left corner of each box shows the sequence of node construction. The root node of the AND-tree corresponds to the OWL class D in Formula \ref{eq:classes}. Since the root is an intersection, the algorithm will handle each operand of it individually through line 4 to 6. The atomic operand $\mathsf{IOInterface}$ is returned directly and added as a child to the root in line 7. For the complex operand $\mathsf{\exists isIEOf.(\geq 3 hasEI.IOInterface)}$, the algorithm recursively generates sub-nodes until the final atomic filler $\mathsf{IOInterface}$ is reached in line 10. Note that all nodes are generated immediately in line 1 when $\mathrm{Construct}$ is called.

It becomes more involved if the OWL class $C$ contains any disjunctions (unions or nominals) because XML does not support \textit{or} statements generally. The solution is to construct $m$ AND-trees for a disjunction with $m$ elements. However, since disjunctions can appear in any nested part inside an OWL class expression, we need to traverse the logical structure of the class expression to produce a set of AND-trees that is logically equivalent to the OWL class.  

\begin{algorithm}[htbp] 
\caption{$\mathrm{ConstructD}$} 
\label{alg:and2} 
\begin{algorithmic}[1] 
\renewcommand{\algorithmicrequire}{\textbf{Input:}}
\renewcommand{\algorithmicensure}{\textbf{Output:}}
\REQUIRE The class expression $ce$ of an OWL class $C$
\ENSURE  A set of tree nodes $roots$

\STATE {initialize $roots = \{\}$}
\IF {$ce$ is an union or a nominal}
    \FOR {each $element$ in $ce$}
        \STATE add ConstructD($element$) to $roots$
    \ENDFOR
\ELSE
    \STATE {make a tree node $n$ for $ce$}, add $n$ to $roots$
    \IF {$ce$ is an atomic class}
        \RETURN $roots$
    \ELSIF {($ce$ is an intersection)}
            \FOR {$operand \in ce$}
                \STATE {let $nestedTrees = $ ConstructD($operand$)}
                \FOR {$root$ in $roots$}
                    \STATE copy $root$ $nestedTrees.size-1$ times
                    \STATE add the root of each $tree \in nestedTrees$ as a successor to exactly one copy of $root$
                \ENDFOR
            \ENDFOR
    \ELSIF {($ce$ is a restriction)}
            \STATE {let $nestedTrees = $ ConstructD($ce.filler$)}
            \FOR {$root$ in $roots$}
                \STATE copy $root$ $nestedTrees.size-1$ times
                \STATE add the root of each $tree \in nestedTrees$ as a successor to exactly one copy of $root$
            \ENDFOR
    \ENDIF
\ENDIF
\RETURN $roots$
\end{algorithmic}
\end{algorithm}

Algorithm \ref{alg:and2} shows the AND-tree construction process for classes involving disjunctions. If the input class expression $ce$ is a disjunction, then a set of tree nodes are generated for the elements of the disjunction (line 4). In case the input is an intersection, the recursive call of $\mathrm{ConstructD}$ in line 12 will handle possible nested disjunction in each element and produce a set of nested trees. These nested trees need to be multiplexed with the existing trees in $roots$ through line 13 to 15. The algorithm treats restrictions similarly to intersections despite that the filler of a restriction is used to produce nested trees in line 19. It is worth noting that only $m-1$ copies of $root$ are made in line 14 and 21 since the original $root$ also counts during the construction.

\begin{figure}[htbp]
\centering
\includegraphics[width=0.5\textwidth]{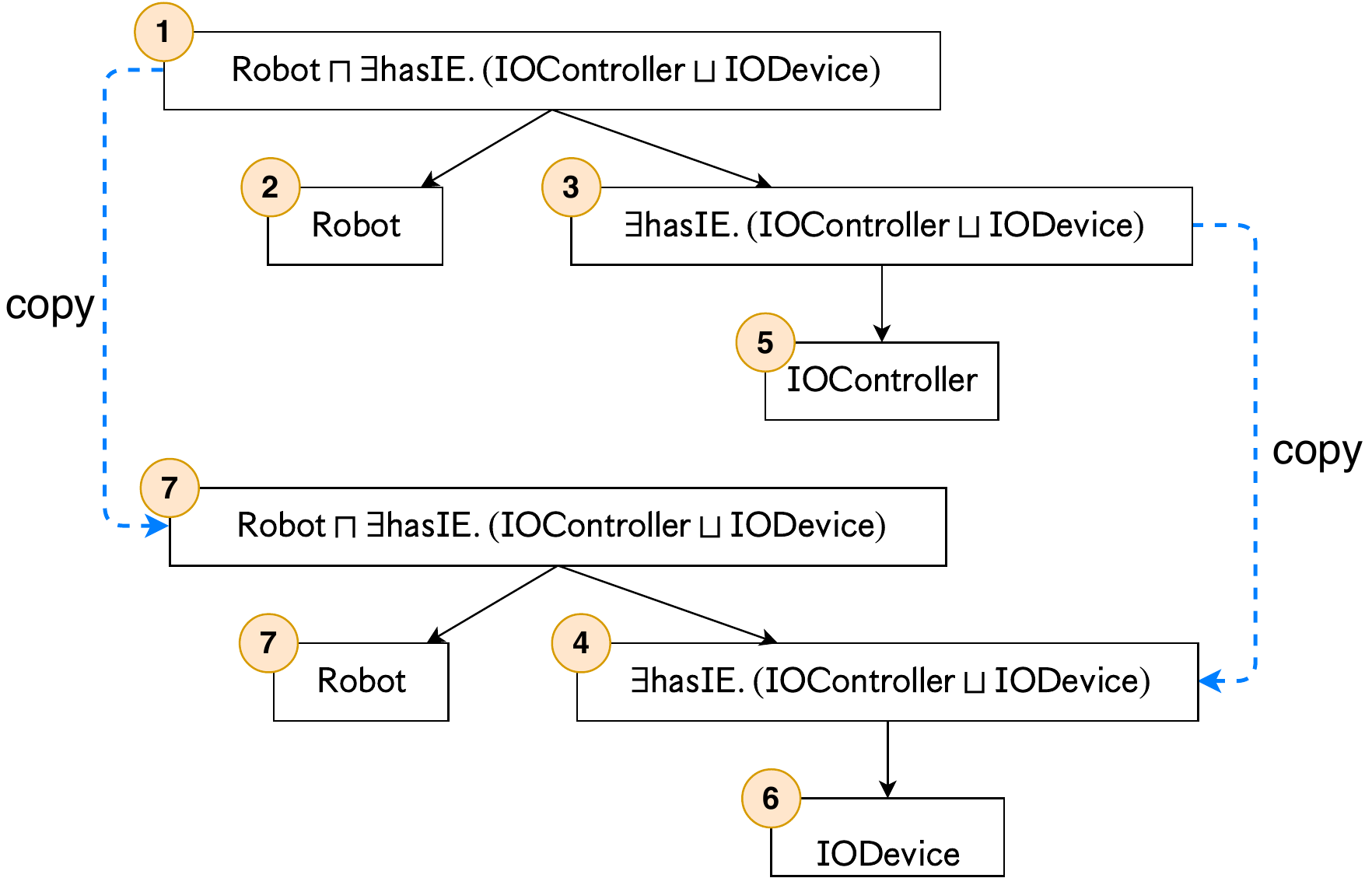}
\caption{The tree construction process of the OWL complex class $\mathsf{Robot \sqcap \exists hasIE.(IOController \sqcup IODevice)}$. The numbers in the tree nodes show the sequence of node construction.}
\label{fig:treeU}
\end{figure}

Figure \ref{fig:treeU} illustrates the tree construction process of the OWL class $\mathsf{Robot \sqcap \exists hasIE.(IOController \sqcup IODevice)}$. In the first step, a root node is generated that contains the complete class expression (line 7). Then, for each operand of the intersection, a child node is generated in step 2 and 3 (line 12). Since the $\mathsf{Robot}$ node is atomic, no further construction is required in the recursive call (line 9). On the other hand, the restriction node $\mathsf{\exists hasIE.(IOController \sqcup IODevice)}$ is copied in step 4 (line 21), since its filler is a union and produces two atomic nodes $\mathsf{IOController}$ and $\mathsf{IODevice}$ (line 19). In step 5 and 6, the atomic nodes are added to the original and copied restriction nodes (line 22). Finally, the root node $\mathsf{Robot \sqcap \exists hasIE.(IOController \sqcup IODevice)}$ is copied once to accept the two distinct restriction nodes in step 7 (line 14-15).

\subsection{Working with Inverse Properties}
\label{sec:inverseRole}

For OWL classes that describe objects in the instance hierarchy, inverse properties might appear for gathering information about their ancestors or siblings (see the OWL classes B and D in Formula \ref{eq:classes}). Due to structural restrictions in AML, we assume that the following conditions hold when an inverse property $R^- \in \mathsf{\{isIEOf, isEIOf\}}$ appear:

\begin{itemize}[itemindent=1em]
    \item[C1:] $R^-$ does not appear in the filler of any restriction that has $R$ as property, e.g. $\exists R.(\exists R^-.C)$.
    \item[C2:] $R^-$ does not appear in the filler of cardinality restrictions, e.g. $\geq n \,\, R^-.C$.
    \item[C3:] $R^-$ does not appear in the filler of any restriction that has a different property $R' \neq R$, e.g. $\exists R'.(\exists R^-.C)$. 
    \item[C4:] $\mathsf{isEIOf}$ does not appear in the filler of any restrictions that has an inverse property, e.g. $\mathsf{\exists R^-.(\exists isEIOf.C)}$
\end{itemize}

\begin{figure*}[htbp]
\centering
\includegraphics[width=1.0\textwidth]{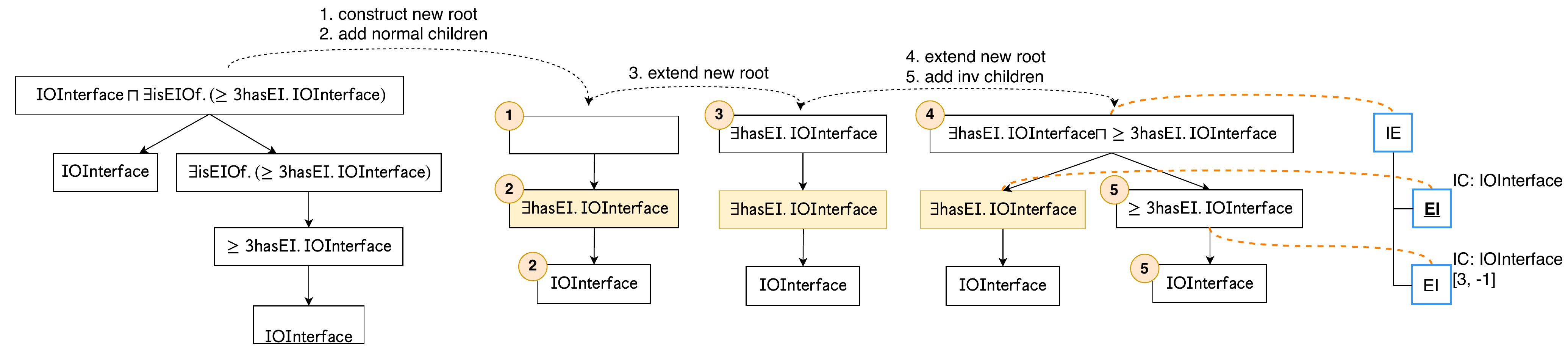}
\caption{The construction of the AML concept tree of class D and the conversion to its AML concept model. The numbers in the tree nodes show the sequence of node construction. The orange dashed lines show the mapping between nodes in the AML concept tree and the AML concept model.}
\label{fig:inverseRoleandAML}
\end{figure*}

The conditions C1 and C2 avoid modeling redundancies in OWL, since AML data has a tree structure, and each node in the tree has a unique predecessor. A class expression $\exists R.\exists R^-.C$ is therefore logically equivalent to $C$, and a cardinality restriction is redundant to an existential restriction. The condition C3 avoids modeling errors in OWL since the set of internal elements is disjoint with the set of external interfaces. The condition C4 holds since external interfaces have no child object in AML. We call an OWL class that meets the conditions C1-C4 as a \textit{proper AML class}. 

The inverse properties of a proper AML class always appear continuously at the outermost layer of the class expression. In other words, the AND-tree of a proper AML class has all inverse properties in the upper part of the tree. Therefore, Algorithm \ref{alg:inverse} iteratively removes the inverse properties from the root of an AND-tree. We call an AND-tree that contains no disjunctions nor inverse properties as an \textit{AML concept tree}. 

\begin{algorithm}[htbp] 
\caption{$\mathrm{removeInverseProperty}$} 
\label{alg:inverse} 
\begin{algorithmic}[1] 
\renewcommand{\algorithmicrequire}{\textbf{Input:}}
\renewcommand{\algorithmicensure}{\textbf{Output:}}
\REQUIRE The root of an AND-tree $root$
\ENSURE The root of a new AND-tree $newRoot$

\STATE {let $ce =$ class expression in $root$}
\IF {$ce$ contains no (nested) inverse property}
    \RETURN {$root$}
\ELSE
    \IF {$ce$ is a restriction}
        \STATE {construct a new node $newRoot$ for $ce.filler$}
        \STATE {change the filler of $root$ to owl:Thing}
        \STATE {add $root$ as a successor of $newRoot$}
        \STATE {move $root.children$ as sucessors of $newRoot$}
    \ELSIF {$ce$ is an intersection}
        \STATE {construct a template node $newRoot$}
        \STATE {let $inv$ be successors of $root$ with inverse property}
        \STATE {let $normal$ be other successors of $root$}
        \STATE {construct a new node $normalChild$ as an existential restriction with $normal$ being its filler}
        \STATE {add $normalChild$ as a successor to $newRoot$}
        \STATE {add the expression of $normalChild$ to $newRoot$}
        \FOR {$node \in inv$}
            \STATE {add the filler of $node$ to $newRoot$ conjunctively}
            \STATE {move $node.child$ as a successor of $newRoot$}
        \ENDFOR
    \ENDIF
    \STATE {RemoveInverseProperty($newRoot$)}
\ENDIF
\end{algorithmic}
\end{algorithm}

Figure \ref{fig:inverseRoleandAML} shows how the inverse property in the root of class D's AND-tree is removed. Since the original root node is an intersection, the algorithm first constructs a template node for the new root (line 11). Then a new child node is constructed for the previous child $\mathsf{IOInterface}$ by formulating an existential restriction in step 2 (line 14 to 15). To keep the consistency of the tree, the expression of the new child node is added to the new root node in the third step (line 16). For the previous child $\mathsf{\exists isEIOf.(\geq 3 hasEI.IOInterface)}$ with the inverse property $\mathsf{isEIOf}$, the filler $\mathsf{\geq 3 hasEI.IOInterface}$ is added to the new root node as a conjunctive term in step 4 (line 18), and the corresponding grandchild with its sub-tree is added as a child to the new root in step 5 (line 19). 

It is obvious that the inverse property $\mathsf{isEIOf}$ is now removed. Note that the OWL class expression of the new root node is different from the original one. Informally, the original root describes the \textit{primary} object in an arbitrary position of the CAEX instance hierarchy (marked as yellow), while the new root describes the predecessor of the primary object.

\subsection{The Forward Translation: from OWL to AML}
\label{sec:tree2aml}

Until now we have shown the algorithms to transform a proper AML class into an AML concept tree. The \textit{forward} translation $\mathrm{TransF: OWL \mapsto AML}$ can be implemented by traversing AML concept trees in a depth-first manner. For every tree node, we generate a corresponding AML concept model whose concept attributes are configured based on the mappings in Table \ref{table:mapping}. The CAEX type of the target AML concept model is determined either by the object property being used in case of a restriction or by the CAEX type annotation of the OWL class in case of an intersection in the root node. The orange dashed lines in Figure \ref{fig:inverseRoleandAML} show the translation from the AML concept tree of class D to its AML concept model illustrated in Figure \ref{fig:classD}. Recall that OWL atomic classes are mapped to CAEX class references.

\subsection{The Backward Translation: from AML to OWL}
\label{sec:aml2owl}

If an AML concept model is proper, i.e. it has exactly one primary element (see section \ref{sec:amlmodel}), then the \textit{backward} translation $\mathrm{TransB: AML \mapsto OWL}$ can be directly carried out using the mappings in Table \ref{table:mapping}. First, a traverse of the AML concept model is necessary to localize the primary object. Afterwards, successors of the primary object are translated to restrictions with normal properties while the predecessors are translated to restrictions with inverse properties. If Algorithm \ref{alg:and2} would have generated several AML concept models during the forward translation, they are translated independently to several OWL classes and combined disjunctively as a union. In this case, an original OWL class with nested unions will be reproduced as a union of expressions, e.g. $\mathsf{\exists r.(C \sqcup D) \to \exists r.C \sqcup \exists r.D}$. 

It is worth noting that the mappings in Table \ref{table:mapping} are used for both $\mathrm{TransF}$ and $\mathrm{TransB}$. Therefore, the forward and backward translation are inverse functions of each other in terms of semantic equivalence. That means, for an OWL class $C$ and an AML concept model $M$, we have:
\begin{align}
\begin{split}
  &\mathrm{TransB(TransF}(C)) \equiv C \\
  &\mathrm{TransF(TransB}(M)) \equiv M
\end{split}
\label{eq:translation}
\end{align}

\section{Use Cases}
\label{sec:usecase}

\begin{figure}[tbp]
\centering
\includegraphics[width=0.35\textwidth]{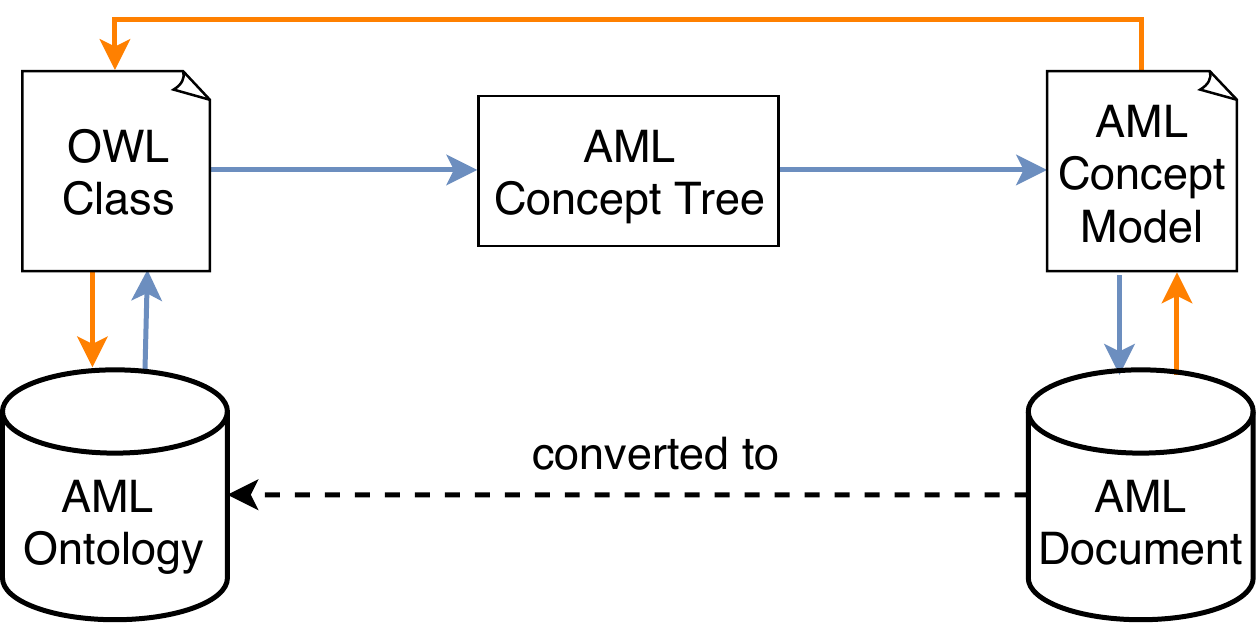}
\caption{The work flow for ontology engineering using bidirectional translation. }
\label{fig:usecase}
\end{figure}

We have implemented the AML concept model and the bidirectional translation in Java. The source code and examples can be found in the GitHub repository\footnote{\url{https://github.com/kit-hua/ETFA2019}}. To demonstrate the use cases of the proposed approach, we discuss two typical scenarios in ontology engineering (Figure \ref{fig:usecase}). For editing AML models, we recommend the AML editor\footnote{\url{https://www.automationml.org/o.red.c/dateien.html}}.

In the first use case (orange arrows in Figure \ref{fig:usecase}), the required OWL class does not exist yet, and a user wants to create an AML concept model for the concept in mind:

\begin{enumerate}
    \item The user generates the primary AML concept model for the target concept, i.e. a CAEX role class, system unit class, interface class, internal element or external interface with class reference and concept attributes.
    \item The user adds CAEX attributes and sub-elements with sufficient constraints to the model. This process repeats recursively for nested attributes and sub-elements.
    \item If the primary AML concept model shall be further restricted by the properties of its predecessor or siblings, a parent AML concept model is generated. This process repeats recursively for further predecessors and siblings.
    \item The user generates the OWL class using the backward translation and adds it to the AML ontology.
\end{enumerate}

The second use case (blue arrows in Figure \ref{fig:usecase}) refers to the ontology evolution procedure in which OWL complex classes already exist in an AML ontology. These classes might be modeled by an ontology expert or created using the AML editor as described above. Now the user might want to inspect a particular OWL class and modify it by demand. First, the chosen OWL class is translated into AML concept models via its AML concept trees. Then, the user can open the generated AML concept models in the AML editor and inspect them by browsing their XML structure. If any modification is necessary, the user can edit the AML concept models as described above and export the new one to an OWL class.

In conclusion, AML concept models can be inspected and modified using the AML editor, while the forward and backward translation are transparent to the user. By comparing the OWL complex classes in Formula \ref{eq:classes} and their corresponding AML concept models in Figure \ref{fig:classes}, we believe that the two use cases demonstrate an intuitive and efficient interaction with OWL complex classes. Because the forward and backward translations are inverse functions of each other (see Formula \ref{eq:translation} in section \ref{sec:aml2owl}), 
Figure \ref{fig:usecase} also illustrates that a round-trip engineering of OWL complex classes is possible by following the work flow of both use cases successively.

\section{Conclusion}
\label{sec:conclussion}

In this paper, we studied the problem of interpreting OWL complex classes from an AML ontology. We identified the inadequacy of existing approaches and introduced a native AML based approach for visualizing, editing and creating OWL complex classes. More specifically, we presented the AML concept model that is capable of carrying ontological semantics, and a bidirectional translation procedure for the conversion between OWL complex classes and AML concept models. With two typical use cases in ontology engineering, we demonstrated the utility of the proposed approach.

Future works are considered in two aspects. First, the semantic expressivity of the AML concept model is restricted by the object properties $\mathsf{hasIE, hasEI}$ and can be extended to cover further modeling facilities in AML, e.g. connections between objects. Second, the current implementation does not provide a friendly user interface and can be improved by integrating the translation procedure into the AML Editor.

\section*{Acknowledgment}

This work has been supported from the European Union’s Horizon 2020 research and innovation programme under grant agreement No 688117 “Safe human-robot interaction in logistic applications for highly flexible warehouses (SafeLog)”.



\bibliographystyle{IEEEtran}
\bibliography{./refs.bib}

\begin{thebibliography}{10}
\providecommand{\url}[1]{#1}
\csname url@samestyle\endcsname
\providecommand{\newblock}{\relax}
\providecommand{\bibinfo}[2]{#2}
\providecommand{\BIBentrySTDinterwordspacing}{\spaceskip=0pt\relax}
\providecommand{\BIBentryALTinterwordstretchfactor}{4}
\providecommand{\BIBentryALTinterwordspacing}{\spaceskip=\fontdimen2\font plus
\BIBentryALTinterwordstretchfactor\fontdimen3\font minus
  \fontdimen4\font\relax}
\providecommand{\BIBforeignlanguage}[2]{{%
\expandafter\ifx\csname l@#1\endcsname\relax
\typeout{** WARNING: IEEEtran.bst: No hyphenation pattern has been}%
\typeout{** loaded for the language `#1'. Using the pattern for}%
\typeout{** the default language instead.}%
\else
\language=\csname l@#1\endcsname
\fi
#2}}
\providecommand{\BIBdecl}{\relax}
\BIBdecl

\bibitem{Rainer2008}
R.~Drath, A.~L\"{u}der, J.~Peschke, and L.~Hundt, ``Automationml - the glue for
  seamless automation engineering,'' in \emph{2008 IEEE International
  Conference on Emerging Technologies and Factory Automation}, Sept 2008, pp.
  616--623.

\bibitem{Hua2018a}
Y.~Hua and B.~Hein, ``Concept {L}earning in {A}utomation{ML} with {F}ormal
  {S}emantics and {I}nductive {L}ogic {P}rogramming,'' in \emph{2018 IEEE
  International Conference on Automation Science and Engineering (CASE)}, 2018.

\bibitem{Abele2013a}
L.~{Abele}, C.~{Legat}, S.~{Grimm}, and A.~W. {Müller}, ``Ontology-based
  {V}alidation of {P}lant {M}odels,'' in \emph{2013 11th IEEE International
  Conference on Industrial Informatics (INDIN)}, July 2013, pp. 236--241.

\bibitem{Hua2016}
Y.~{Hua}, S.~{Zander}, M.~{Bordignon}, and B.~{Hein}, ``From {A}utomationml to
  {ROS}: {A} {M}odel-driven {A}pproach for {S}oftware {E}ngineering of
  {I}ndustrial {R}obotics using {O}ntological {R}easoning,'' in \emph{2016 IEEE
  21st International Conference on Emerging Technologies and Factory Automation
  (ETFA)}, Sep. 2016, pp. 1--8.

\bibitem{Hildebrandt2018}
C.~{Hildebrandt}, S.~{Törsleff}, B.~{Caesar}, and A.~{Fay}, ``Ontology
  {B}uilding for {C}yber-{P}hysical {S}ystems: {A} domain expert-centric
  approach,'' in \emph{2018 IEEE 14th International Conference on Automation
  Science and Engineering (CASE)}, Aug 2018, pp. 1079--1086.

\bibitem{Runde09}
S.~Runde, K.~G{\"u}ttel, and A.~Fay, ``Transformation von
  {CAEX}-{A}nlagenplanungsdaten in {OWL}: {E}ine {A}nwendung von {T}echnologien
  des {S}emantic {W}eb,'' in \emph{Automation 2009, Der
  Automatisierungskongress in Deutschland, VDI/VDE-Gesellschaft Mess- und
  Automatisierungstechnik (GMA)}, Jun 2009, pp. 175--178.

\bibitem{Kovalenko2015}
O.~{Kovalenko}, M.~{Wimmer}, M.~{Sabou}, A.~{Lüder}, F.~J. {Ekaputra}, and
  S.~{Biffl}, ``Modeling {A}utomation{ML}: {S}emantic {W}eb technologies vs.
  {M}odel-{D}riven {E}ngineering,'' in \emph{2015 IEEE 20th Conference on
  Emerging Technologies Factory Automation (ETFA)}, Sep. 2015, pp. 1--4.

\bibitem{Runde10}
S.~Runde, A.~Fay, and S.~B{\"o}hm, ``{K}onvertierung von
  {OWL}-{P}lanungsergebnissen nach {CAEX},'' in \emph{Automation 2010, Der 11.
  Branchentreff der Mess- und Automatisierungstechnik, VDI/VDE-Gesellschaft
  Mess- und Automatisierungstechnik (GMA)}, Jun 2010, pp. 1--12.

\bibitem{OWL2DL}
\BIBentryALTinterwordspacing
B.~Motik, P.~F. Patel-Schneider, and B.~C. Grau, ``{OWL} 2 {W}eb {O}ntology
  {L}anguage {D}irect {S}emantics ({S}econd {E}dition),'' 11.12.2012. [Online].
  Available: \url{https://www.w3.org/TR/owl2-direct-semantics/}
\BIBentrySTDinterwordspacing

\bibitem{Markus2012}
\BIBentryALTinterwordspacing
M.~Kr{\"{o}}tzsch, F.~Simancik, and I.~Horrocks, ``A {D}escription {L}ogic
  {P}rimer,'' \emph{CoRR}, vol. abs/1201.4089, 2012. [Online]. Available:
  \url{http://arxiv.org/abs/1201.4089}
\BIBentrySTDinterwordspacing

\bibitem{OWL2}
\BIBentryALTinterwordspacing
B.~Motik, P.~F. Patel-Schneider, and B.~Parsia, ``{OWL} 2 {W}eb {O}ntology
  {L}anguage {D}irect {S}emantics ({S}econd {E}dition),'' 11.12.2012. [Online].
  Available: \url{https://www.w3.org/TR/owl2-syntax}
\BIBentrySTDinterwordspacing

\bibitem{Hitzler:2009}
P.~Hitzler, M.~Krtzsch, and S.~Rudolph, \emph{Foundations of Semantic Web
  Technologies}, 1st~ed.\hskip 1em plus 0.5em minus 0.4em\relax Chapman \&
  Hall/CRC, 2009.

\end{thebibliography}
%



\end{document}